\pdfoutput=1
\documentclass[conference]{IEEEtran}
\IEEEoverridecommandlockouts
\usepackage{cite}
\usepackage{amsmath,amssymb,amsfonts}
\usepackage{algorithmic}
\usepackage{graphicx}
\usepackage{textcomp}
\def\BibTeX{{\rm B\kern-.05em{\sc i\kern-.025em b}\kern-.08em
    T\kern-.1667em\lower.7ex\hbox{E}\kern-.125emX}}

\usepackage[table,xcdraw]{xcolor}
\usepackage{colortbl, booktabs}
\usepackage{algorithm}
\usepackage{algorithmic}
\usepackage{amsfonts}
\usepackage{amsmath}
\usepackage{makecell}
\usepackage{multirow}
\usepackage{booktabs}
\usepackage{arydshln}
\usepackage{amssymb}
\usepackage{pifont}

\newcommand{\CUT}[1]

\begin{document}

\title{Uneven Event Modeling for Partially Relevant Video Retrieval}


\author{Sa Zhu$^{1,2,3}$ \quad Huashan Chen$^{1*}$ \quad Wanqian Zhang$^{1*}$ \thanks{*Corresponding author} \quad  Jinchao Zhang$^{1,3}$ \\ \quad  Zexian Yang$^{1,2}$ \quad Xiaoshuai Hao$^4$  \quad Bo Li$^{1,3}$
\\ Institute of Information Engineering,
Chinese Academy of Sciences$^1$ \\ School of Cyber Security, University of Chinese Academy of Sciences
$^2$ \\  State Key Laboratory of Cyberspace Security Defense$^{3}$ \quad Beijing Academy of Artificial Intelligence$^4$
\\ \{zhusa, chenhuashan, zhangwanqian, zhangjinchao, yangzexian, libo\}@iie.ac.cn
}

\maketitle

\begin{abstract}
Given a text query, partially relevant video retrieval (PRVR) aims to retrieve untrimmed videos containing relevant moments, wherein event modeling is crucial for partitioning the video into smaller temporal events that partially correspond to the text.
Previous methods typically segment videos into a fixed number of equal-length clips, resulting in ambiguous event boundaries. 
Additionally, they rely on mean pooling to compute event representations, inevitably introducing undesired misalignment.
To address these, we propose an Uneven Event Modeling (UEM) framework for PRVR. 
We first introduce the Progressive-Grouped Video Segmentation (PGVS) module, to iteratively formulate events in light of both temporal dependencies and semantic similarity between consecutive frames, enabling clear event boundaries. 
Furthermore, we also propose the Context-Aware Event Refinement (CAER) module to refine the event representation conditioned the text's cross-attention.
This enables event representations to focus on the most relevant frames for a given text, facilitating more precise text-video alignment. 
Extensive experiments demonstrate that our method achieves state-of-the-art performance on two PRVR benchmarks. Code is available at https://github.com/Sasa77777779/UEM.git.

\end{abstract}


\section{Introduction}
\label{sec:intro}

With the rapid expansion of online video content, text-to-video retrieval (T2VR)~\cite{hao2023dual,hao2023uncertainty, fang2023uatvr, wang2023unified, wang2024text, yang2024dgl,  jin2023text,hao2021multi,hao2021matters} has gained significant research attention. 
Traditional T2VR methods typically assume that videos are 
pre-trimmed to short durations,
enabling precise semantic alignment with the query.
However, on platforms such as YouTube, videos are often \textit{untrimmed}, comprising multiple events, and a single textual caption can represent only a fragment of the overall content. 
This divergence highlights a critical gap between conventional T2VR research and its real-world applications.


Recently, a novel Partially Relevant Video Retrieval (PRVR) task has been developed~\cite{dong2022partially}, leveraging untrimmed video databases.
In particular, \textit{a single video may correspond to multiple relevant queries, while each query only details one specific event within the associated video.} 
For PRVR task, \textbf{event modeling} is identified as its primary challenge, which aims at dividing the video into smaller, coherent temporal units capturing specific events that may be partially relevant to the given query, as discussed in~\cite{wang2024gmmformer-v2, wang2024gmmformer}. 
Generally, to construct events for partial text matching, existing PRVR methods first segment consecutive frames into equal-sized clips with a fixed number, each of which is regarded as an event. 
Then, the event representation is computed by applying mean pooling to its corresponding frame features. 
Finally, the text-video similarity is determined by the alignment between the text features and the event-level video features. 

\begin{figure}[!t]
  \centering
  \setlength{\abovecaptionskip}{-0.1em}
  \includegraphics[width=1.0\linewidth]{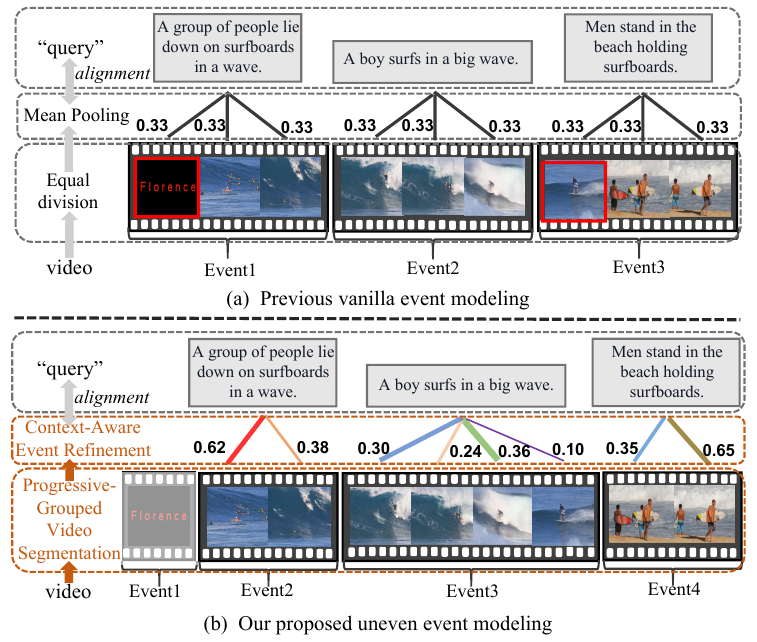}
  \caption{
  Two paradigms for event modeling in PRVR task:
  (a) Previous methods segment videos into a fixed number of equal-length events and employ mean pooling to compute event representations, which results in ambiguous event boundaries, e.g., between Event 2 and 3. (b) Our uneven event modeling adaptively groups frames into multiple varying-length events, and computes the event representation based on the attention weights of the given text, achieving clearer event boundaries and more precise text-video alignment.
  }
  \label{Motivation}
  \vspace{-1em}
\end{figure}

Although this event modeling paradigm has achieved impressive results, it still suffers from two problems: 1) Both the number and duration of events depicted in each video vary significantly. 
Directly segmenting videos into a fixed number of equal-length clips often leads to ambiguities in event boundaries, causing the confused alignment between video events and texts.
For example, as shown in Fig.~\ref{Motivation} (a), frames corresponding to the same event may be dispersed across multiple clips, while adjacent frames representing distinct events may be erroneously grouped within the same clip. 
2) Within an event, the relevance of each frame to the given text also varies significantly.
Trivially utilizing mean pooling to aggregate frame features for event representation may inadvertently encode superfluous or even distracting visual information that is unrelated to the text, thereby potentially degrading the retrieval performance.

To address the above-mentioned two issues, in this paper, we propose an Uneven Event Modeling (UEM) framework to adaptively cluster and aggregate video frames for event modeling in the PRVR task. 
Specifically, we first introduce the Progressive-Grouped Video Segmentation (PGVS) module to dynamically segment the videos into multiple events. 
Unlike previous methods that divide all videos into a fixed number of equal-sized clips, PGVS forms event clusters iteratively by evaluating the similarity between consecutive frames.
In this temporally progressive manner, PGVS assigns each frame to the same event if they are semantically similar.
Thus it can adaptively construct events with varying numbers and durations for each video, leading to clear event boundaries regarding both semantic and temporal information. 
Next, we propose the Context-Aware Event Refinement (CAER) module to compute event representations by adaptively aggregating associated frames, conditioned on the attention weights of the given text. 
Compared to trivially mean pooling for text-agnostic event representation, our CAER module allows an event to reason about the most relevant frames to a given text, facilitating more precise alignment.
We summarize our main contributions as follows:
\begin{itemize}
  \item We design an Uneven Event Modeling (UEM) framework that adaptively clusters and aggregates video frames for effective event modeling in the PRVR task.
  \item We propose the Progressive-Grouped Video Segmentation (PGVS) module, which could adaptively construct events with varying numbers and durations for each video, leading to clearer event boundaries. 
  \item To further enable the event representation to focus on the most relevant frames for a given text, we propose the Context-Aware Event Refinement (CAER) module, which refines the event representation conditioned on the attention weights of the given text, facilitating more precise alignment.
  \item Extensive experiments on two widely used benchmark datasets demonstrate that our proposed method achieves state-of-the-art performance on the PRVR task. 
\end{itemize}

\section{Related Work}
\textbf{Partially Relevant Video Retrieval (PRVR)} aims to retrieve untrimmed videos that are partially relevant to a given query~\cite{dong2022partially}. 
Previous PRVR methods either focus on modeling the correlations between different clips, such as \cite{dong2022partially}, which applies a multi-scale sliding window strategy to explicitly interact clip embeddings, or \cite{wang2024gmmformer}, which introduces a Gaussian-Mixture Block (GMMFormer) to implicitly model the relationships between consecutive clips. 
Alternatively, other approaches investigate text-clip matching, for example, \cite{wang2024gmmformer-v2}, which treats the assignment of texts and events as a maximum matching problem and introduces an optimal matching loss for fine-grained alignment between text queries and relevant clips.
Although they have demonstrated great success, they typically construct events by dividing all frames into a fixed number of equal-length clips and compute event representations by mean pooling associated frames. 
However, this leads to ambiguities in event boundaries, even encoding extraneous information that is unrelated to the text query.

\textbf{Event Modeling} typically employs two paradigms for video segmentation: Equal Division and K-means Clustering. Equal Division, widely used in PRVR tasks~\cite{dong2023dual,wang2024gmmformer,wang2024gmmformer-v2,diao2024tasar,jiang2023progressive}, divides frames into fixed-length clips sequentially. While efficient, it ignores frame similarities, resulting in unclear event boundaries. K-means Clustering~\cite{wang2024videotree} treats frames as independent instances, clustering them into 
$k$ groups. Although it groups similar frames, it neglects temporal dependencies within events. Both methods require predefining the number of clusters, limiting their adaptability to videos with varying event counts and durations.


\section{Methodology}

\subsection{Problem Formulation}
\begin{figure*}[!t]
\centering
  \setlength{\abovecaptionskip}{-0.1em}
  \includegraphics[width=0.97\textwidth]{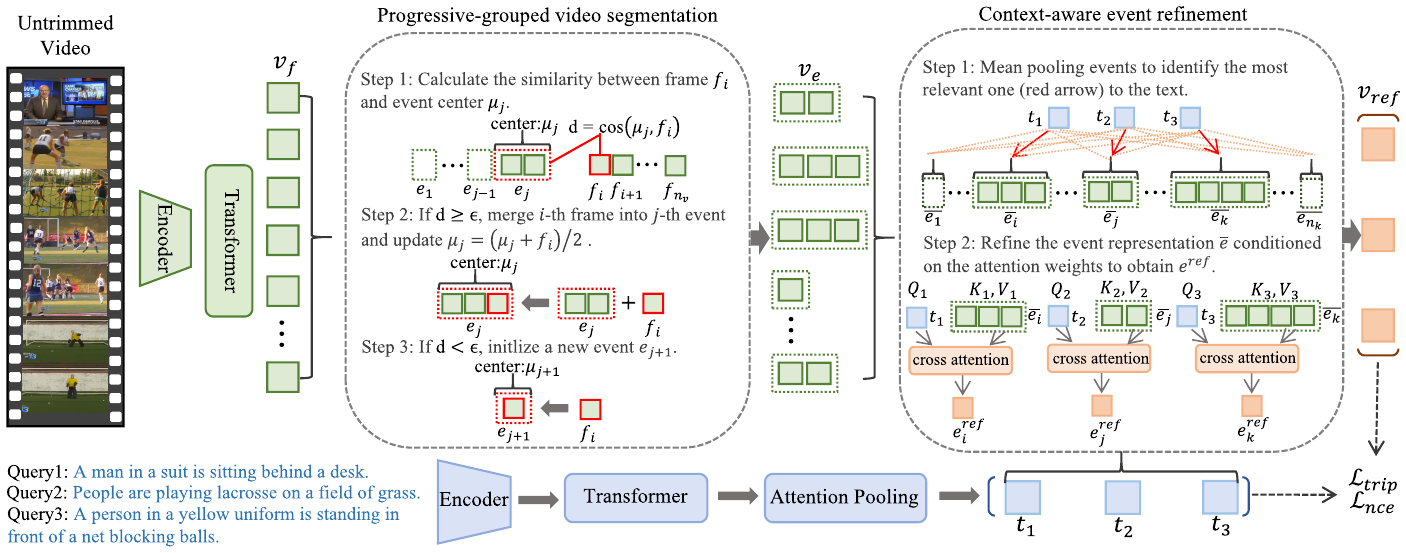}
  \caption{An overview of the proposed Uneven Event Modeling framework, which consists of the Progressive-Grouped Video Segmentation (PGVS) module and the Context-Aware Event Refinement (CAER) module. Specifically, the PGVS module progressively groups frames into multiple events utilizing temporal dependencies between consecutive frames via the similarity threshold $\epsilon$, while the CAER module adaptively aggregates frames conditioned on the text for event representation computation.
  }
 \label{framework}
 \vspace{-0.5em}
\end{figure*}

Given a video-text dataset consisting of untrimmed videos $\mathcal{V}$ and texts $\mathcal{T}$, where each video $v_{i} \in \mathcal{V}$ contains multiple events and is associated with several text descriptions, and each text description $t_{i} \in \mathcal{T}$ corresponds to the content of a specific event in the associated video, the Partially Relevant Video Retrieval (PRVR) task aims to retrieve the video that contains the event semantically relevant to a given query $t^{q}$ from a large video database. 
To achieve this objective, we propose an Uneven Event Modeling (UEM) framework, as illustrated in Fig.~\ref{framework}, which primarily comprises three parts: text query encoding, video event modeling, and partial-relevant text-video matching. 
For event modeling, we introduce two modules: the Progressive-Grouped Video Segmentation (PGVS) module and the Context-Aware Event Refinement (CAER) module, designed for adaptive video segmentation and event representation computation, respectively.

\subsection{Text Query Encoding}
Given a sentence consisting of $n_t$ words, we first input it into CLIP's text encoder to obtain a sequence of word token embeddings. Then we employ a fully connected (FC) layer with ReLU activation to project the word embeddings into a lower-dimensional space. Subsequently, we add learnable positional embeddings to the projected word embeddings and process them through a Transformer encoder layer to obtain a sequence of contextualized word embedding vectors $Q = \{q_{i}\}_{i=1}^{n_t}$, where $q_{i} \in \mathbb{R}^{d}$ denotes the $i$-th word feature and $d$ is the feature dimension.
Finally, following ~\cite{dong2022partially}, we apply a simple attention pooling mechanism on $Q$ to obtain sentence embeddings $t \in \mathbb{R}^{d}$ as follows:
\begin{equation}
  t=\sum_{i=1}^{n_t}a_{i}^{q}q_{i},\  a^{q}=softmax(\omega Q^{T}),
\end{equation}
where $\omega \in \mathbb{R}^{d}$ is a trainable vector and $a^{q} \in \mathbb{R}^{1 \times n_t}$ indicates the attention vector.


\subsection{Video Event Modeling}

\noindent
\textbf{Progressive-Grouped Video Segmentation}.
To segment videos into multiple events, existing PRVR methods divide all frames into a fixed number of equal-length clips. 
However, as discussed above, this segmentation approach is sub-optimal for untrimmed videos with varying event number and duration, as it will cause the ambiguities in event boundaries. 
The K-means algorithm can also be used for frame clustering~\cite{wang2024videotree}, but it not only neglects the temporal dependencies between consecutive frames, but also requires the number of clusters to be pre-defined.
To this end, we introduce the Progressive-Grouped Video Segmentation (PGVS) module to adaptively construct events. 

Specifically, given a video containing $n_v$ frames, we first extract visual features for each frame with the pre-trained CLIP visual encoder, 
obtaining a sequence of frame features $v_F \in \mathbb{R}^{n_v \times d_v}$. 
Then, following~\cite{dong2022partially}, we employ a standard Transformer with a learned positional embedding to improve the temporal dependency of the frame features:
\begin{equation}
  v_f = \{f_i\}_{i=1}^{n_v} = Transformer(FC(v_F) + PE),
\end{equation}
where $PE$ denotes the positional embeddings, $v_f \in \mathbb{R}^{d}$ represents the frame features for each video.

\begin{algorithm}
    \caption{Progressive-Grouped Video Segmentation}
    \renewcommand{\algorithmicrequire}{\textbf{Input:}}
    \renewcommand{\algorithmicensure}{\textbf{Output:}}
    \label{Sequence Clustering} 
    \begin{algorithmic}[1]
        \REQUIRE video frames $v_f = \{f_i\}_{i=1}^{n_v}$; similarity threshold $\epsilon$; 
        \ENSURE Event cluster $v_e = \{e_j\}_{j=1}^{n_k}$;
        \STATE Initialize video clusters $v_e = \{\}$; event centers $\mu$;
        \FOR{frame $f_i$ in $v_f$}
        \IF {$i=1$}
        \STATE Set $e_1=\{f_1\}$, $\mu=f_1$;
        \ELSE
        \STATE Calculate the similarity $d$ between frame $f_i$ and cluster center $\mu$;
        \IF {$d \geq \epsilon$}
        \STATE $e_j \leftarrow e_j \cup \{f_i\}$;
        \STATE Update $\mu = (\mu+f_i)/2$;
        \ELSE
        \STATE $e_{j+1}=\{f_i\}$; $v_e \leftarrow e_j \cup e_{j+1}$;
        \STATE Update $\mu=f_i$;
        \ENDIF
        \ENDIF
        \ENDFOR
            
    \end{algorithmic}  
\end{algorithm}

After that, as shown in Algorithm~\ref{Sequence Clustering}, we progressively group the video frames into multiple events. 
The process begins with the initialization of event clusters and their corresponding centers, followed by the iterative assignment of frames based on a pre-defined similarity threshold. 
Specifically, if the similarity between a frame and the current event center exceeds the threshold, the frame is assigned to the corresponding event cluster. 
Otherwise, a new event cluster is created. 
After the assignment, the event center is re-calculated to reflect the frames within the cluster. 
This iterative procedure continues until all frames are assigned. 
The similarity between a frame $f_i$ and a cluster center $\mu$ is computed as follows:
\begin{equation}
  s(f_i, \mu) = \frac{f_i \cdot \mu}{\lVert f_i \rVert \lVert \mu \rVert}.
\end{equation}

Compared to other clustering algorithms, our progressive group algorithm does not require the cluster number to be pre-defined. 
Instead, it relies on a pre-defined similarity threshold, where video frames are clustered iteratively based on this threshold. 
Therefore, it can adaptively construct events with varying numbers and durations while preserving the temporal dependencies of adjacent frames within each event, leading to clearer event boundaries.

\noindent
\textbf{Context-Aware Event Refinement}. 
After segmenting the videos into multiple events, we proceed to explore the computation of event representations. 
Since the relevance of each frame within an event to the given text varies significantly, we introduce a Context-Aware Event Refinement (CAER) module. 
CAER leverages the cross-attention mechanism to adaptively aggregate frames, refining the event representation to enhance its sensitivity to semantically relevant frames while suppressing irrelevant ones.
Specifically, for each video, we first apply mean pooling to the frame-level features within each event to obtain the coarse-grained event representations $\bar{e}$. 
Then, given a query, we compute its similarity to each event and select the most relevant one, denoted as $e^{max}=\{f_i\}_{i=1}^{n_{e^{max}}}$. 
After that, we project the text embedding $t \in \mathbb{R}^{d}$ into query $Q_t \in \mathbb{R}^{1 \times D_p}$ and the frame embeddings within selected event 
into key $K_e \in \mathbb{R}^{n_{e^{max}}\times D_p}$ and value $V_e \in \mathbb{R}^{ n_{e^{max}}\times D_p}$ matrices, where $D_p$ is the size of the projection dimension. 
The projections are defined as:
\begin{align}
Q_t &= LN(t)W_Q, \\
K_e &= LN(e^{max})W_K, \\
V_e &= LN(e^{max})W_V, 
\end{align}
where $LN$ is a Layer Normalization layer and $W_Q$, $W_K$
and $W_V$ are projection matrices in $\mathbb{R}^{d \times D_p}$. 
Finally, we adapt the cross-attention from the text embedding to the frame embeddings to obtain the text-refined event representation as:
\begin{equation}
  e^{ref} = MLP(softmax(\frac{Q_{t}K_{e}^{T}}{\sqrt{D_p}})V_e).
\end{equation}

Since the attention weights capture the relevance of a given text to each frame within an event, the refined event representation contains less redundant semantic information, resulting in more precise alignment.

\subsection{Partial-relevant Text-video Matching}
The text-video similarity is measured as the cosine similarity between the refined event representation $e^{ref} \in \mathbb{R}^{1 \times d}$ and the query feature $t\in \mathbb{R}^{1 \times d}$ as follows:
\begin{equation}
  S(v,q) = cos(e^{ref}, t).
\end{equation}

We jointly adopt triplet ranking loss~\cite{faghri2017improving} and infoNCE loss~\cite{zhang2021video}, that are widely used in retrieving related tasks, to optimize the model.
Given a positive video-text pair $(v,q)$, the triplet ranking loss over the mini-batch $\mathcal{B}$ is defined as:
\begin{multline}
  \mathcal{L}_{trip} = \frac{1}{n} \sum_{(v,q) \in \mathcal{B}} \{max(0,m+S(v, q^{-})-S(v,q)) \\ 
  + max(0,m+S(v^{-}, q)-S(v,q))\},
\end{multline}
where $m$ is a margin constant, $q^{-}$ and $v^{-}$ indicate a negative
text for $v$ and a negative video for $q$. 
Similar to~\cite{dong2022partially}, we randomly sample the negative samples from the mini-batch at the beginning of the training and choose the hardest negative samples after 20 epochs.
The infoNCE loss is computed as:
\begin{multline}
  \mathcal{L}_{nce} = -\frac{1}{n} \sum_{(v,q) \in \mathcal{B}} \{log(\frac{S(v,q)}{S(v,q)+\sum_{q_{i}^{-} \in \mathcal{N}_q}S(v, q_i^-)}) \\ 
  + log(\frac{S(v,q)}{S(v,q)+\sum_{v_{i}^{-} \in \mathcal{N}_v}S(v_i^-, q)})  \},
\end{multline}
where $\mathcal{N}_q$ denotes all negative texts of the video $v$ in the mini-batch, while $\mathcal{N}_v$ denotes all negative videos of the text $q$ in the mini-batch.

Finally, the overall objective can be formulated as:
\begin{equation}
   \mathcal{L} = \mathcal{L}_{trip} + \lambda \mathcal{L}_{nce},
\end{equation}
where $\lambda$ serves as the hyper-parameter for balancing.




\section{Experiments}
\subsection{Experimental Setting}
\noindent
\textbf{Datasets.}
We evaluate the performance of our method on two benchmark datasets: 
ActivityNet Captions and TV show Retrieval (TVR). 
\textbf{ActivityNet Captions}~\cite{krishna2017dense} was originally developed for dense video captioning task, and now it has been re-purposed for partially relevant video retrieval. 
It comprises approximately 20K videos from YouTube and the average length of videos is around 118 seconds. 
Each video has around 3.7 moments paired with a corresponding descriptive sentence. 
For a fair comparison, we adopt the same data partition used in ~\cite{dong2023dual}. 
\textbf{TVR}~\cite{lei2020tvr} is a multimodal dataset originally designed for video corpus moment retrieval.  
It contains 21.8K videos collected from 6 TV shows, with an average duration of approximately 76 seconds per video. 
Five sentences are associated with each video, describing different moments in the video. 
We utilize the same data partition in ~\cite{wang2024gmmformer, dong2022partially}.

\noindent
\textbf{Baselines.}
We conduct comparisons against three \textit{text-to-video retrieval (T2VR) models}, including DE~\cite{dong2019dual}, W2VV++~\cite{li2019w2vv++} and Cap4Video~\cite{wu2023cap4video}, three \textit{video corpus moment retrieval (VCMR) models}, including XML~\cite{lei2020tvr}, ReLoCLNet~\cite{zhang2021video} and CONQUER~\cite{hou2021conquer}, and five \textit{partially relevant video retrieval (PRVR) models}, including MS-SL~\cite{dong2022partially}, PEAN~\cite{jiang2023progressive}, DL-DKD~\cite{dong2023dual}, GMMFormer~\cite{wang2024gmmformer} and GMMFormer-v2~\cite{wang2024gmmformer-v2}. 
For VCMR models, the training process consists of two stage: firstly retrieving candidate videos, then localizing specific moments within these videos.
Since moment annotations are unavailable in PRVR, following~\cite{dong2022partially, wang2024gmmformer}, we re-trained the VCMR models by removing the moment localization stage.

\noindent
\textbf{Evaluation Metrics.}
Following the previous works~\cite{wang2024gmmformer, dong2022partially,hao2022listen,hao2023mixgen,HAO2025103018}, we utilize rank-based metrics, namely R@K (K = 1, 5, 10, 100). 
R@K denotes the proportion of queries that successfully retrieve the desired items within the top K of the ranking list. 
The SumR is also utilized as the overall performance. 
Higher scores indicate better performance.

\noindent
\textbf{Implementation Details.}
For video representation, we utilize the frame feature provided by~\cite{dong2023dual}, that is, 512-D visual feature obtained by the CLIP’s ViT-B/32 image encoder. 
For sentence representation, we employ CLIP's transformer-based text encoder to extract 512-D textual features for each word.  
For model training, we set the initial learning rate to 0.0002 and use the same learning schedule as ~\cite{lei2020tvr}. The maximum number of
epochs is set to 100 and batch size is set to 64. we empirically set the hyper parameter $\gamma$ in Eq.11 to 0.02.

\begin{table}
\renewcommand{\arraystretch}{1.3}
  \caption{Performance compared with the state-of-the-arts on activitynet-captions.}
  \setlength{\abovecaptionskip}{-0.1em}
  
  \resizebox{1.0\linewidth}{!}{
  \begin{tabular}{c|ccccc}
   \hline
     Method & R@1 & R@5 & R@10 & R@100 & SumR \\
    \hline
    DE~\cite{dong2019dual}(CVPR'19) & 5.6 & 18.8 & 29.4 & 67.8 & 121.7 \\
    W2VV++~\cite{li2019w2vv++}(MM'19) & 5.4 & 18.7 & 29.7 & 68.8 & 122.6 \\
    Cap4Video~\cite{wu2023cap4video}(CVPR'23) & 6.3 & 20.4 & 30.9 & 72.6 & 130.2\\
    \hline
    XML~\cite{lei2020tvr}(ECCV'20) & 5.3 & 19.4 & 30.6 & 73.1 & 128.4 \\
     ReLoCLNet~\cite{zhang2021video}(SIGIR'21) & 5.7 & 18.9 & 30.0 & 72.0 & 126.6 \\
    CONQUER~\cite{hou2021conquer}(MM'21) & 6.5 & 20.4 & 31.8 & 74.3 & 133.1 \\
    \hline
    MS-SL~\cite{dong2022partially}(MM'22) & 7.1 & 22.5 & 34.7 & 75.8 & 140.1 \\
    PEAN~\cite{jiang2023progressive}(ICME'23) & 7.4 & 23.0 & 35.5 & 75.9 & 141.8 \\
    DL-DKD~\cite{dong2023dual}(CVPR'23) & 8.0 & 25.0 & 37.5 & 77.1 & 147.6 \\
    GMMFormer~\cite{wang2024gmmformer}(AAAI'24) & 8.3 & 24.9 & 36.7 & 76.1 & 146.0 \\
    GMMFormer-v2~\cite{wang2024gmmformer-v2}(arxiv'24) & \underline{8.9} & \underline{27.1} & \underline{40.2} & \underline{78.7} & \underline{154.9} \\
    \rowcolor{gray!50}\textbf{UEM (Ours)} & \textbf{11.8} & \textbf{32.0} & \textbf{45.2} & \textbf{82.2} & \textbf{171.2} \\
    \hline
  \end{tabular}
  }
 \label{act}
 \vspace{-0.5em}
\end{table}

\begin{table}
\renewcommand{\arraystretch}{1.3}
  \caption{Performance compared with the state-of-the-arts on TVR.}
  
  \resizebox{1.0\linewidth}{!}{
  \begin{tabular}{c|ccccc}
   \hline
    Method & R@1 & R@5 & R@10 & R@100 & SumR \\
    \hline
    DE~\cite{dong2019dual}(CVPR'19) & 7.6 & 20.1 & 28.1 & 67.6 & 123.4 \\
    W2VV++~\cite{li2019w2vv++}(MM'19) & 5.0 & 14.7 & 21.7 & 61.8 & 103.2 \\
    Cap4Video~\cite{wu2023cap4video}(CVPR'23) & 10.3 & 26.4 & 36.8 & 74.0 & 147.5\\
    \hline
    XML~\cite{lei2020tvr}(ECCV'20) & 10.0 & 26.5 & 37.3 & 81.3 & 155.1 \\
     ReLoCLNet~\cite{zhang2021video}(SIGIR'21) & 10.7 & 28.1 & 38.1 & 80.3 & 157.1 \\
    CONQUER~\cite{hou2021conquer}(MM'21) & 11.0 & 28.9 & 39.6 & 81.3 & 160.8 \\
    \hline
    MS-SL~\cite{dong2022partially}(MM'22) & 13.5 & 32.1 & 43.4 & 83.4 & 172.4 \\
    PEAN~\cite{jiang2023progressive}(ICME'23) & 13.5 & 32.8 & 44.1 & 83.9 & 174.2 \\
    DL-DKD~\cite{dong2023dual}(CVPR'23) & 14.4 & 34.9 & 45.8 & 84.9 & 179.9 \\
    GMMFormer~\cite{wang2024gmmformer}(AAAI'24) & 13.9 & 33.3 & 44.5 & 84.9 & 176.6 \\
    GMMFormer-v2~\cite{wang2024gmmformer-v2}(arxiv'24) & \underline{16.2} & \underline{37.6} & \underline{48.8} & \underline{86.4} & \underline{189.1} \\
    \rowcolor{gray!50}\textbf{UEM (Ours)} & \textbf{24.4} & \textbf{49.2} & \textbf{60.5} & \textbf{91.3} & \textbf{225.4} \\
    \hline
  \end{tabular}
  }
 \label{tvr}
 \vspace{-0.5em}
\end{table}

\subsection{Comparisons with State-of-the-arts}
Tables~\ref{act} and~\ref{tvr} present the comparison of the retrieval performance between various baselines and our proposed UEM model on the ActivityNet Captions and TVR datasets, respectively. 
Generally, PRVR models show better performance than T2VR and VCMR models, which can be attributed to their ability to model the partial relevance between videos and texts. 
Besides, our UEM model surpasses all previous works by a significant margin. 
Specifically, it achieves a 10.5\% and 19.6\% relative lift in SumR over the past SOTA competitor, GMMFormer-v2, on two benchmarks, respectively. 
This indicates that: 1) the PGVS module dynamically groups frames into multiple events, enabling more flexible event construction, and 2) the CAER module computes event representations conditioned on the given text, leading to more concise text-video alignment.

\subsection{Ablation Studies}
\noindent
\textbf{The effect of different components.} 
To understand the effect of each component, we ablate two main components of our proposed UEM framework, including the Progressive-Grouped Video Segmentation (PGVS) module and the Context-Aware Event Refinement (CAER) module.
As shown in Table~\ref{Ablation on components}, both components effectively improve the model performance. 
In particular, the comparison between No.1 and No.2 demonstrates that the PGVS module yields a substantial improvement of 12.9\% in $R@1$. 
Experiment of No.4 outperforms that of No.1 by a clear margin, demonstrating that the PGVS and CAER modules can work synergistically to model event representation more precisely.

\begin{table}
\renewcommand{\arraystretch}{1.3}
  \caption{Ablation study to investigate the effect of different components on activitynet-captions.}
  \centering
  \resizebox{0.9\linewidth}{!}{
  \begin{tabular}{c|cc|ccccc}
   \hline
    No. & PGVS & CAER & R@1 & R@5 & R@10 & R@100 & SumR \\
    \hline
   1 & \textbf{\ding{56}} & \textbf{\ding{56}} & 
   8.9 & 26.8 & 39.9 & 78.9 & 154.5 \\
   2 & \textbf{\ding{52}} & \textbf{\ding{56}} & 11.4 & 31.0 & 44.2 & 80.9 & 167.5\\
   3 & \textbf{\ding{56}} & \textbf{\ding{52}} & 11.3 & 31.5 & 44.9 & 81.9 & 169.6\\
    \rowcolor{gray!50} 4 &\textbf{\ding{52}} & \textbf{\ding{52}} & \textbf{11.8} & \textbf{32.0} & \textbf{45.2} & \textbf{82.2} & \textbf{171.2} \\
    \hline
  \end{tabular}
  }
 \label{Ablation on components}
 \vspace{-0.5em}
\end{table}

\begin{table}
\renewcommand{\arraystretch}{1.3}
  \caption{Ablation study to investigate the effect of different event modeling methods on activitynet-captions.}
   \centering
  \resizebox{0.9\linewidth}{!}{
  \begin{tabular}{c|ccccc}
   \hline
    Method & R@1 & R@5 & R@10 & R@100 & SumR \\
    \hline
   Equal division & 10.1 & 29.0 & 41.9 & 80.6 & 161.6 \\
   K-means cluster & 10.7 & 30.2 & 42.1 & 80.3 & 163.3 \\
    \rowcolor{gray!50} \textbf{PGVS} & \textbf{11.4} & \textbf{31.0} & \textbf{44.2} & \textbf{80.9} & \textbf{167.5} \\
    \hline
  \end{tabular}
  }
 \label{Ablation on event construction method}
 \vspace{-0.5em}
\end{table}

\noindent
\textbf{Different event modeling methods.} 
We also compare PGVS module with two variants: \textit{Equal division} which segments videos into 32 equal-length events~\cite{dong2022partially}, and \textit{K-means clustering} which utilizes the K-means algorithm to assign frames to the nearest center of 32 clusters.
As in Table~\ref{Ablation on event construction method}, PGVS module outperforms both variants. 
Obviously, PGVS not only considers frame similarities but also accounts for the temporal dependencies between frames during event construction, resulting in more concise text-video alignment.

\begin{figure}[h]
\centering
\setlength{\abovecaptionskip}{-0.1em}
\includegraphics[width=\linewidth]{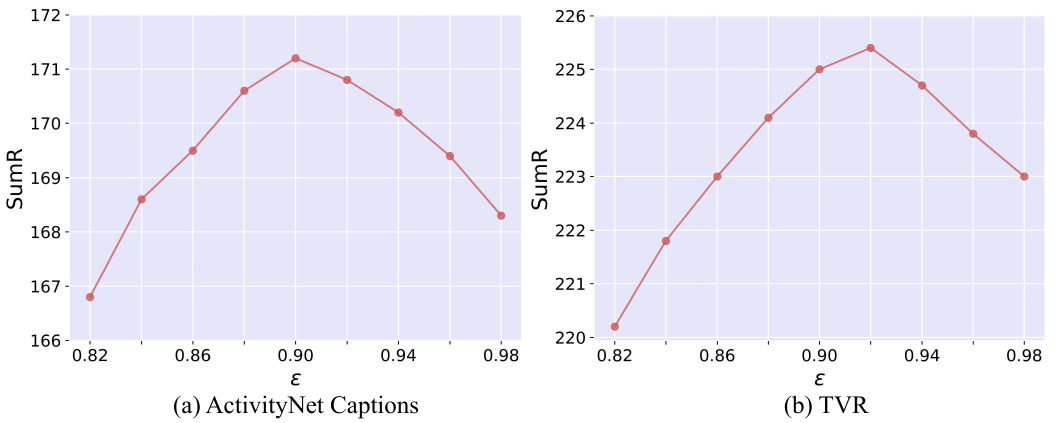}
\caption{Analysis of the similarity threshold $\epsilon$ in PGVS module.}
\label{threshold analysis}
\vspace{-0.5em}
\end{figure}

\noindent
\textbf{Analysis of similarity threshold in PGVS.} 
We further investigate the impact of the similarity threshold $\epsilon$ on retrieval performance. 
As shown in Fig.~\ref{threshold analysis}, as the threshold $\epsilon$ increases, SumR initially rises and then declines. 
This can be attributed to the following factors: when the threshold is small, frames that describe the same event are divided across multiple segments, leading to incomplete information within each segment. 
On the contrary, a larger similarity threshold results in longer event durations, which may introduce irrelevant information. 
Thus, we set $\epsilon$ to 0.90 and 0,92 for the ActivityNet Captions and TVR datasets, respectively.

\subsection{Qualitative Analysis}
\begin{figure}[t]
\centering
\setlength{\abovecaptionskip}{-0.1em}
\includegraphics[width=0.95\linewidth]{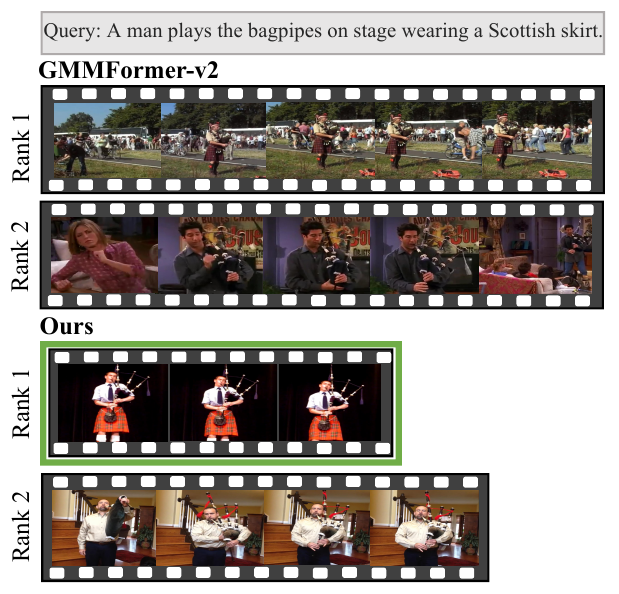}
\caption{Top-2 text-to-video retrieval results on ActivityNet Captions. The target event is displayed with the green box.}
\label{quantization analysis}
\vspace{-1em}
\end{figure}

Fig.~\ref{quantization analysis} shows the qualitative retrieval results of our method on ActivityNet Captions dataset.
It is evident that our UEM retrieves more relevant video events in response to the query text compared to the SOTA method GMMFormer-v2. 
Additionally, the events constructed by UEM exhibit clearer boundaries, highlighting the effectiveness of our PGVS module.

\section{Conclusion}
In this paper, we introduce UEM, an Uneven Event Modeling framework for PRVR. 
Specifically, we first propose the Progressive-Grouped Video Segmentation (PGVS) module, which adaptively constructs events with varying numbers and durations for each video, resulting in clear event boundaries that capture both semantic and temporal information. 
Furthermore, we present the Context-aware Event Refinement (CAER) module, which refines the event representation to emphasize the most relevant frames for a given text, thereby enabling more precise alignment. 
Extensive experiments and ablation studies on two widely used benchmarks validate the effectiveness of our proposed method for the PRVR task.

\section{Acknowledgment}
This work was supported by the National Natural Science Foundation of China under Grants 62202459, the National Key R\&D Program of China under Grant 2022YFB3103500, and the Open Research Project of the StateKey Laboratory of Industrial Control Technology, China (Grant No.ICT2024B51).

\bibliographystyle{IEEEbib}
\bibliography{main}


\end{document}